%% file: admpo.tex
\documentclass{article}

\usepackage{arxiv}

\usepackage[utf8]{inputenc} 
\usepackage[T1]{fontenc}    
\usepackage{hyperref}       
\usepackage{url}            
\usepackage{booktabs}       
\usepackage{amsfonts}       
\usepackage{nicefrac}       
\usepackage{microtype}      
\usepackage{xcolor}         
\usepackage{amsmath}
\usepackage{algorithm}
\usepackage{algorithmic}
\usepackage{multirow}

\usepackage[normalem]{ulem}

\usepackage{amsthm}
\theoremstyle{plain}
\newtheorem{theorem}{Theorem}[section]

\theoremstyle{definition}
\newtheorem{definition}[theorem]{Definition}
\newtheorem{assumption}[theorem]{Assumption}

\usepackage{subfigure}
\usepackage{graphicx}

\title{Any-step Dynamics Model Improves Future Predictions for Online and Offline Reinforcement Learning}
\author{%
    Haoxin Lin\textsuperscript{\rm 1,\rm 2,\rm 3},~~
    Yu-Yan Xu\textsuperscript{\rm 3},~~
    Yihao Sun\textsuperscript{\rm 1,\rm 2},~~
    Zhilong Zhang\textsuperscript{\rm 1,\rm 2,\rm 3},~~
    Yi-Chen Li\textsuperscript{\rm 1,\rm 2,\rm 3},~~ \\
    \textbf{
    Chengxing Jia\textsuperscript{\rm 1,\rm 2,\rm 3},~~
    Junyin Ye\textsuperscript{\rm 1,\rm 2,\rm 3},~~
    Jiaji Zhang\textsuperscript{\rm 1,\rm 2},~~
    Yang Yu\textsuperscript{\rm 1,\rm 2,\rm 3,\rm 4}\thanks{Corresponding Author}} \\   
    \textsuperscript{\rm 1}National Key Laboratory for Novel Software Technology, Nanjing University, Nanjing, China\\
    \textsuperscript{\rm 2}School of Artificial Intelligence, Nanjing University, Nanjing, China\\
    \textsuperscript{\rm 3}Polixir Technologies, Nanjing, China\\
    \textsuperscript{\rm 4}Peng Cheng Laboratory, Shenzhen, 518055, China\\
    linhx@lamda.nju.edu.cn, yuyan.xu@polixir.ai,\\ \{sunyh, zhangzl, liyc, jiacx, yejy, zhangjj\}@lamda.nju.edu.cn, yuy@nju.edu.cn
}
\date{}

\begin{document}

\maketitle

\begin{abstract}
Model-based methods in reinforcement learning offer a promising approach to enhance data efficiency by facilitating policy exploration within a dynamics model. However, accurately predicting sequential steps in the dynamics model remains a challenge due to the bootstrapping prediction, which attributes the next state to the prediction of the current state. This leads to accumulated errors during model roll-out. In this paper, we propose the \textbf{A}ny-step \textbf{D}ynamics \textbf{M}odel (ADM) to mitigate the compounding error by reducing bootstrapping prediction to direct prediction. ADM allows for the use of variable-length plans as inputs for predicting future states without frequent bootstrapping. We design two algorithms, ADMPO-ON and ADMPO-OFF, which apply ADM in online and offline model-based frameworks, respectively. In the online setting, ADMPO-ON demonstrates improved sample efficiency compared to previous state-of-the-art methods. In the offline setting, ADMPO-OFF not only demonstrates superior performance compared to recent state-of-the-art offline approaches but also offers better quantification of model uncertainty using only a single ADM.
\end{abstract}

\section{Introduction}
Model-based Reinforcement Learning (MBRL) \cite{mbrl} has demonstrated empirical success in both online \cite{mve, steve, pets, slbo, mbpo, mppve} and offline \cite{mopo, morel, combo, rambo, mobile, morec} settings. The essence of MBRL lies in the dynamics model, where extensive explorations and evaluations of the agent can occur, thereby reducing the reliance on real-world samples. Embedded in the model-based framework, online policy optimization can leverage a large Update-To-Data (UTD) ratio \cite{redq} to improve sample efficiency, while offline policy optimization can be completed using the model augmented data beyond the dataset.

Although some efforts aim to propose high-fidelity dynamics models, such as adversarial models \cite{advm,armor}, causal models \cite{zhu2022arxiv}, and ensemble dynamics models \cite{pets, mbpo, mopo} adopted by the majority of MBRL algorithms, it is challenging to generate high-quality imaginary samples via long-horizon model roll-out. In a dynamics model with the common form, the state-action pair at time step $t$, $(s_t,a_t)$, is used as input to predict the next state $s_{t+1}$. Thus, the bootstrapping prediction, which attributes the next state to the prediction of the current state, is inevitably employed to roll out states in the dynamics model. The deviation error of generated states increases with the roll-out length since the error accumulates gradually as the state transitions in imagination. If updated on the unreliable samples with a large compounding error, the policy will be misled by biased policy gradients.

In the online setting, the impact of compounding error \cite{xu2021error} on policy optimization restricts the utilization of the model, thereby hindering further improvement in sample efficiency. In the offline setting, compounding error affects the accuracy of current model uncertainty estimation based on ensemble. For instance, using the behavior policy corresponding to the dataset for a long-horizon roll-out in the model still leads to a great compounding error, and the accumulated deviations of different learners cannot be similar. In this case, the divergence of the ensemble will inevitably be large, which is inconsistent with the situation that the actual trajectory is within the region covered by the dataset. Therefore, it is essential to reduce compounding error in both online and offline settings.

One potential way to deal with the issue of compounding error is to reduce bootstrapping prediction to direct prediction, considering the direct state transition after executing a multi-step action sequence \cite{msmbrl, m3, combining, temp_abs}. Although state $s_{t+1}$ is only dependent on state-action $(s_t,a_t)$ under the assumption of Markov property \cite{rl}, the prediction of $s_{t+1}$ can actually leverage earlier information. Tracing back a prior $k$-step plan, $s_{t+1-k}$ and the intermediate $k$-step actions $(a_{t+1-k},a_{t+2-k},\cdots,a_{t})$ are sufficient to constitute an attribution to predict $s_{t+1}$. 

To handle the variable-length plans, we introduce a special \textbf{A}ny-step \textbf{D}ynamics \textbf{M}odel (ADM) that allows for the use of $s_{t+1-k}$ and $(a_{t+1-k},a_{t+2-k},\cdots,a_{t})$ corresponding to any integer $k$ within a specified range as inputs for predicting $s_{t+1}$. When the agent faces changes occurring in the trajectory distribution, the state predictions from different backtracking lengths will exhibit noticeable divergence. This feature naturally enables ADM to estimate model uncertainty without ensemble. Replacing the ensemble dynamics model with ADM, we devise a unique model roll-out method with random backtracking, which can be plugged into any existing MBRL algorithmic frameworks. In this paper, our main purpose is to demonstrate how the augmented data generated by ADM exhibits excellent effectiveness, both in improving future predictions and measuring the model uncertainty.

In general, our contributions are summarized as follows. (1) We present a generalized dynamics model called ADM to replace the dynamics model used in existing online and offline MBRL algorithms and demonstrate its superiority in reducing the compounding error. (2) We propose a new online MBRL algorithm called ADMPO-ON based on ADM and show that it can outperform recent state-of-the-art online model-based algorithms in terms of sample efficiency while retaining competitive performance on MuJoCo \cite{mujoco} benchmarks. (3) We propose a new offline MBRL algorithm called ADMPO-OFF based on ADM and show that it can effectively quantify the model uncertainty, achieving superior performance compared to recent state-of-the-art offline algorithms on D4RL \cite{d4rl} and NeoRL \cite{neorl} benchmarks.

\section{Preliminaries}
\subsection{Markov Decision Process and Reinforcement Learning}
We consider a standard Markov Decision Process (MDP) specified by a tuple $\mathcal{M}=(\mathcal{S}, \mathcal{A}, T, \rho_0, \gamma)$, where $\mathcal{S}$ is the state space, $\mathcal{A}$ is the action space, $T(s_{t+1},r_{t+1}|s_t,a_t)$ is the dynamics function that calculates the conditioned distribution of $s_{t+1}\in\mathcal{S}$ and $r_{t+1}\in\mathbb{R}$ given $(s_t, a_t)$, $\rho_0$ is the initial state distribution, and $\gamma$ is the discount factor. We use $\rho^\pi$ to denote the on-policy distribution over states induced by the dynamic function $T$ and the policy $\pi$. From a multi-step perspective, the attribution of state $s_{t+1}$ and reward $r_{t+1}$ can be traced back to the earlier $k$-step plan, $s_{t-k+1}$ along with the action sequence $a_{t-k+1:t}=(a_{t-k+1},a_{t-k+2},\cdots,a_t)$ in between. This relationship can be represented by the $k$-step dynamics model
\begin{equation}
T^k(s_{t+1},r_{t+1}|s_{t-k+1},a_{t-k+1:t})=\sum_{(s_{t-k+2:t},r_{t-k+2:t})}\prod_{i=0}^{k-1} T(s_{t-i+1},r_{t-i+1}|s_{t-i},a_{t-i}).
\end{equation}
We use $\Gamma^k_\pi(s_{t-k+1:t},a_{t-k+1:t}|s_{t+1})$ to denote the distribution over $(s_{t-k+1:t},a_{t-k+1:t})$ conditioned on $s_{t+1}$ induced by the dynamic function $T$ and the policy $\pi$.

The optimization goal of Reinforcement Learning (RL) is to find a policy $\pi$ that maximized the expected discounted return $\mathbb{E}_{\rho^\pi}\left[\sum_{t=1}^\infty \gamma^{t-1} r_t\right]$. Such a policy can be derived from the estimation of the state-action value function
$Q^\pi(s_t,a_t)=\mathbb{E}_{(s_{t+1},r_{t+1})\sim T(\cdot|s_t,a_t)}\left[r_{t+1}+\gamma V^\pi(s_{t+1})\right]$, where $V^\pi(s_{t+1})=\mathbb{E}_{a_{t+1}\sim\pi(\cdot|s_{t+1})}\left[Q^\pi(s_{t+1},a_{t+1})\right]$ is the state value function.

\subsection{Model-based Reinforcement Learning}
MBRL aims to find the optimal policy while transferring the agent's explorations and evaluations from the environment to the learned dynamics model. Given a dataset $\mathcal{D}_{\mathrm{env}}$ collected via interaction in the real environment, the dynamics model $\hat{T}$ is typically trained to maximize the expected likelihood $\mathbb{E}_{(s_t,a_t,r_{t+1},s_{t+1})\sim\mathcal{D}_{\mathrm{env}}}[\log\hat{T}(s_{t+1},r_{t+1}|s_t,a_t)]$. The estimated dynamics model defines a surrogate MDP $\hat{\mathcal{M}}=(\mathcal{S}, \mathcal{A}, \hat{T}, \rho_0, \gamma)$. Then any RL algorithm can be used to recover the optimal policy with the augmented dataset $\mathcal{D}_{\mathrm{env}}\cup\mathcal{D}_{\mathrm{model}}$, where $\mathcal{D}_{\mathrm{model}}$ is the synthetic data rolled out in $\hat{\mathcal{M}}$.

The above-mentioned paradigm is adopted by model-based policy optimization (MBPO) \cite{mbpo} and much of its follow-up work \cite{mppve,ddppo,m2ac,maac} in the online setting. These works don't need to consider the issue of model coverage, as the agent can explore online to fill in the regions where the dynamics model is uncertain. However, in the offline setting, the limited dataset causes $\hat{T}$ to cover only a part of the state-action space. Once the agent encounters out-of-distribution samples during roll-out in $\hat{\mathcal{M}}$, the learning process can collapse. Therefore, MOPO \cite{mopo} and some of its subsequent offline MBRL algorithms \cite{morel, mobile} incorporate a penalty term in the reward function to measure the model uncertainty, allowing the agent to sample within safe regions of $\hat{T}$.

\section{Method}
In this section, we propose a special \textbf{A}ny-step \textbf{D}ynamics \textbf{M}odel (ADM) to replace the mainstream ensemble dynamics models. Applying ADM to existing MBRL frameworks for policy optimization, we introduce two algorithms, namely online ADMPO-ON and offline ADMPO-OFF. ADM improves future state predictions since it reduces bootstrapping prediction to direct prediction by backtracking variable-length plans. Consequently, ADMPO-ON can improve the sample efficiency in the online setting, while ADMPO-OFF can accurately estimate the model uncertainty in the offline setting.

\subsection{Any-step Dynamics Model}
Currently, the prevalent dynamics models typically operate on a single-step basis, with $s_t$ and $a_t$ as inputs to predict $s_{t+1}$ and $r_{t+1}$. In a broader context, dynamics models can also be multi-step \cite{msmbrl, m3, combining}, where inputs encompass $s_t$ along with a $k$-step sequence of actions $(a_t,a_{t+1},\cdots,a_{t+k-1})$ to predict $s_{t+k}$ and $r_{t+k}$. To introduce flexibility in the backtracking length of the model, we further extend the definition of the multi-step dynamics model to allow $k$ to be any positive integer within a specified range, as delineated in Definition \ref{any_step_model}.

\begin{definition}[Any-step Dynamics Model]
\label{any_step_model}
    Given the maximum backtracking length $m$, an any-step dynamics model $\hat{T}(s_{t+k},r_{t+k}|s_t,a_{t:t+k-1})$ is the distribution of $s_{t+k}\in\mathcal{S}$ and $r_{t+k}\in\mathbb{R}$ conditioned on the $k$-step plan $(s_t, a_{t:t+k-1})=(s_t, a_t,a_{t+1},\cdots,a_{t+k-1})\in\mathcal{S}\times\mathcal{A}^k$, where $k$ can be any integer between $[1,m]$.
\end{definition}

\begin{figure*}[pt!]
    \centering
    \includegraphics[width=\linewidth]{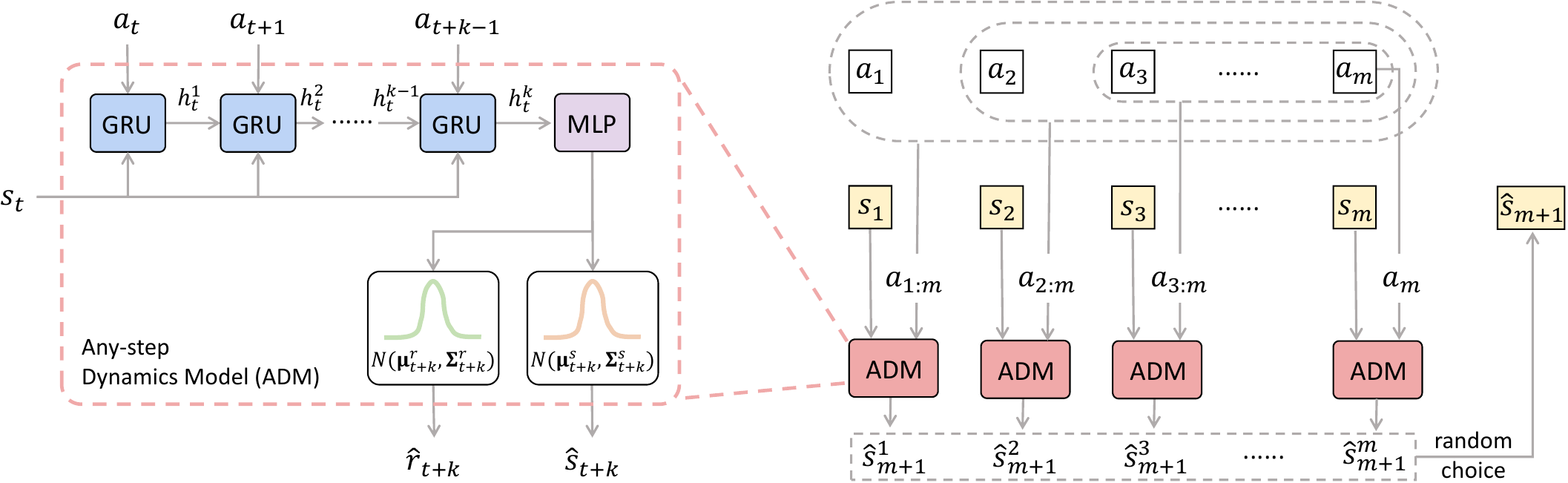}
    \caption{Illustration of any-step dynamics model (left) structured using RNN and its application for next-step prediction with random backtracking (right).}
    \label{ADM}
\end{figure*}

To handle inputs with variable step sizes, we utilize an RNN \cite{rnn} with a GRU \cite{gru} cell to implement the any-step dynamics model, as depicted in the left part of Figure \ref{ADM}. Certainly, Transformer \cite{transformer} is also a feasible choice, but we do not consider it because the model structure is beyond the scope of this study. Since the input state consists of only one step, while the action may be a sequence of multiple steps, we duplicate the state to match the length of the action sequence, and then sequentially feed it into the RNN. The input $(s_t,a_{t:t+k-1})$, after being represented by the RNN, yields the hidden $h^k_t$, which is then fed into an MLP to obtain the mean and standard deviation of $s_{t+k}$ and $r_{t+k}$, i.e., $(\boldsymbol{\mu}^s_{t+k}, \boldsymbol{\Sigma}^s_{t+k})$ and $(\boldsymbol{\mu}^r_{t+k}, \boldsymbol{\Sigma}^r_{t+k})$. Similar to previous model-based methods \cite{mbpo, m2ac, mppve}, we model the distributions of $s_{t+k}$ and $r_{t+k}$ as Gaussian distributions and predict them through sampling. We call the \textbf{A}ny-step \textbf{D}ynamics \textbf{M}odel as ADM and denote it as $\hat{T}_\theta(s_{t+k},r_{t+k}|s_t,a_{t:t+k-1})$, where $\theta$ represents the neural parameters. With the real samples from the environment, $\hat{T}_\theta$ is trained to maximize the expected likelihood:
\begin{equation}
    \label{model_obj}
    J_T(\theta)=\frac{1}{m}\sum_{k=1}^m\mathbb{E}_{(s_t,a_{t:t+k-1},r_{t+k},s_{t+k})\sim\mathcal{D}_{\textrm{env}}}
    \left[\log \hat{T}_\theta(s_{t+k},r_{t+k}|s_t,a_{t:t+k-1})\right].
\end{equation}

With $\hat{T}_\theta$, the frequent bootstrapping during model roll-out can be reduced. Specifically, given the maximum backtracking length $m$, a state-action sequence of length $m$, $(s_1,a_1,s_2,a_2,\cdots,s_m,a_m)$, is sampled from the data buffer to start the roll-out in $\hat{T}_\theta$. For predicting $s_{m+1}$, an integer between $[1,m]$ is chosen uniformly at random as the backtracking length. If only one step is selected for backtracking, $(s_m,a_m)$ will be fed into $\hat{T}_\theta$ to obtain the prediction result; if $m-1$ steps are chosen, $(s_2,a_{2:m})$ will be fed into $\hat{T}_\theta$, and so forth. The right part of Figure \ref{ADM} illustrates the aforementioned process based on random backtracking. Further prediction for $s_{m+2}$ should backtrack to at most $(s_2,a_{2:m+1})$, as $\hat{T}_\theta$ is trained with a maximum sequence length of $m$ steps. Similarly, subsequent state predictions can also backtrack at most $m$ steps. The backtracked state, one part of the attribution for next state prediction, is located several steps ahead in expectation. Thus, ADM reduces the actual bootstrapping count of a rolled-out trajectory. The complete $H$-step roll-out process in ADM is described in Algorithm \ref{ADM_rollout}.

\begin{algorithm}[t!]
    \caption{Roll-out in ADM: \textbf{ADM-Roll}($\hat{T}_\theta$, $\pi_\phi$, $H$, $m$, $(s_1,a_1,s_2,a_2,\cdots,s_{m-1},a_{m-1},s_m)$)}
    \label{ADM_rollout}
    \textbf{Input}: Learned ADM $\hat{T}_\theta$ with parameters $\theta$, policy $\pi_\phi$ with parameters $\phi$, roll-out length $H$, maximum backtracking length $m$, state-action sequence $(s_1,a_1,s_2,a_2,\cdots,s_{m-1},a_{m-1},s_m)$

    \begin{algorithmic}[1]
        \FOR{$\tau=0$ to $H-1$}
        \STATE Sample $a_{m+\tau}\sim\pi_\phi(\cdot|s_{m+\tau})$
        \STATE Randomly sample an integer $k$ from $[1,m]$ uniformly
        \STATE Roll out the next step in ADM via $(s_{m+\tau+1},r_{m+\tau+1})\sim\hat{T}_\theta(\cdot|s_{m+\tau+1-k},a_{m+\tau+1-k:m+\tau})$
        \ENDFOR
        \RETURN $(s_m, a_m, r_{m+1}, s_{m+1},\cdots,s_{m+H-1},a_{m+H-1},r_{m+H},s_{m+H})$
    \end{algorithmic}
\end{algorithm}

Similar to existing MBRL algorithms, policy roll-out in ADM can generate a large number of fake samples for policy updates. We refer to the new dyna-style ADM-based policy optimization framework as ADMPO (\textbf{ADM}-based \textbf{P}olicy \textbf{O}ptimization). Any policy optimization algorithm can be plugged into this framework. In the subsequent subsections, we will introduce two foundational algorithms, AMRPO-ON and ADMPO-OFF, for online and offline settings, respectively.

\subsection{ADMPO-ON: ADM for Policy Optimization in Online Setting}
In the online setting, the agent interacts with the real environment while simultaneously optimizing the policy. Like MBPO \cite{mbpo}, ADMPO-ON can be divided into two alternating stages, namely updating the dynamics model with continuously collected samples and utilizing samples generated through model roll-outs additionally for policy optimization. ADMPO-ON replaces the ensemble dynamics model in MBPO framework with ADM. It trains ADM with the optimization objective shown in Equation \eqref{model_obj} and generates a large number of fake samples using the roll-out method depicted in Algorithm \ref{ADM_rollout}. The detailed pseudo-code is provided by Algorithm \ref{admpo_on_code} in Appendix \ref{admpo_on}.

During roll-outs, ADM randomly selects a backtracking length at each step and attributes the states to be predicted to variable-length plans. While backtracking $k$ steps, we view the sampling process $(\hat{s}_{t+1},\hat{r}_{t+1})\sim\hat{T}_\theta(\cdot|s_{t-k+1},a_{t-k+1:t})$ as $(\hat{s}_{t+1},\hat{r}_{t+1})=\mu_\theta(s_{t-k+1},a_{t-k+1:t})+\eta_{t+1}$ with $\eta_{t+1}\sim \mathcal{N}(0,\Sigma_\theta(s_{t-k+1},a_{t-k+1:t}))$, where $\mu_\theta$ is the deterministic dynamics function and $\Sigma_\theta$ is the standard deviation function used to construct the noise distribution with zero mean. In expectation, the target value of $Q(s_{t},a_{t})$ is estimated as
\begin{equation}
\label{q_trgt}
\mathbb{E}_{(s_{t-m+1:t-1},a_{t-m+1:t-1})\sim\Gamma^{m-1}_\pi(\cdot|s_t)}\left[\frac{1}{m}\sum_{k=1}^m\mathbb{E}_{(\hat{s}_{t+1},\hat{r}_{t+1})\sim \hat{T}_\theta(\cdot|s_{t-k+1},a_{t-k+1:t})}
\left[y(\hat{s}_{t+1},\hat{r}_{t+1})\right]\right],
\end{equation}
where $y(\hat{s}_{t+1},\hat{r}_{t+1})=\hat{r}_{t+1}+\gamma \mathbb{E}_{a\sim\pi(\cdot|\hat{s}_{t+1})}\left[Q(\hat{s}_{t+1},a)\right]$. Data generated via roll-outs in our ADM can be viewed as an implicit augmentation. The augmentation stems from two sources: (i) variation of the backtracking-length while applying the learned ADM to predict the next state, and (ii) the noise introduced by the distribution $\mathcal{N}(0,\Sigma_\theta(s_{t-k+1},a_{t-k+1:t}))$ at each backtracking-length $k$. According to \cite{ensemble_necessary}, variations of state predictions can effectively implicitly regularize the local Lipschitz condition of the Q network around regions where the model prediction is uncertain, thereby regulating the value-aware model error \cite{value_aware_model_error}. 

\subsection{ADMPO-OFF: ADM for Policy Optimization in Offline Setting}
In the offline setting, due to limitations of the behavior policy corresponding to the dataset, the learned ADM can only cover some regions of the state-action space. Beyond these safe regions lie the risky regions where the model is uncertain and unable to be fixed since online exploration is inaccessible to the agent. To prevent policy optimization collapse, exploitation of the learned model needs to be focused within the safe regions. Simultaneously, efforts should be made to explore beyond the boundaries of the risky regions to discover samples conducive to a better policy than the behavior policy. Achieving such a balance between conservatism and generalization often requires measuring model uncertainty. Based on ADM, next we will introduce a new uncertainty quantification method.

In our ADM, states predicted using different backtracking lengths exhibit discrepancies. Intuitively, these discrepancies are closely related to the data distribution. When the agent is in safe regions, the discrepancies are small. As the agent gradually explores towards risky regions, the discrepancies tend to increase. The difference among probabilistic predictions $\hat{T}_\theta(\cdot|s_{t-k+1},a_{t-k+1:t})$ obtained with different backtracking $k$ serve as a natural measure of model uncertainty, which can be quantified using variance (or standard deviation), as defined by Definition \ref{ADM_uncertainty_def}.

\begin{definition}[ADM-Uncertainty Quantifier]
\label{ADM_uncertainty_def}
For any maximum backtracking length $m$ and the corresponding learned ADM $\hat{T}_\theta$, the uncertainty of $\hat{T}_\theta$ at $(s_t,a_t)$ is quantified as
\begin{equation}
\label{ADM_uncertainty}
\begin{aligned}
\mathcal{U}^\mathrm{ADM}(s_t,a_t)=&\mathbb{E}_{\Gamma^{m-1}_\pi(\cdot|s_t)}
\left[\left\|\mathrm{Var}_{k\sim \mathrm{Uniform}(m),\hat{s}_{t+1}\sim \hat{T}_\theta(\cdot|s_{t-k+1},a_{t-k+1:t})}\left[\hat{s}_{t+1}\right]\right\|_1\right]\\
=&\mathbb{E}_{\Gamma^{m-1}_\pi(\cdot|s_t)}
\left[\left\|\frac{1}{m}\sum_{k=1}^m\left((\Sigma_\theta^k)^2+(\mu_\theta^k)^2\right)-(\bar{\mu})^2\right\|_1\right]
\end{aligned}
\end{equation}
for any $s_t\in\mathcal{S}$ and $a_t\in\mathcal{A}$, where $\Sigma_\theta^k=\Sigma_\theta(s_{t-k+1},a_{t-k+1:t})$, $\mu_\theta^k=\mu_\theta(s_{t-k+1},a_{t-k+1:t})$ for convenience, and $\bar{\mu}=\frac{1}{m}\sum_{k=1}^m \mu_\theta^k$.
\end{definition}

This uncertainty term corresponds to a combination of epistemic and aleatoric model uncertainty with a similar form to the ensemble standard deviation \cite{offm_design, ensemble_uncertainty}. However, the source of diversity has shifted from ensemble to variable backtracking lengths. Since estimating the approximation error via epistemic or aleatoric uncertainty has been applied in many works \cite{mopo,pbrl,mobile,offm_design}, we assume that our ADM uncertainty \eqref{ADM_uncertainty} is an admissible error estimator \cite{mopo}, as described in Assumption \ref{admissible_error}.
\begin{assumption}[Admissible Error Estimator]
\label{admissible_error}
Assume that there exists a positive $b\in\mathbb{R}^+$ such that the following inequality \eqref{admissible_error_cond} holds for any maximum backtracking length $m$ and any $s_t\in\mathcal{S}$, $a_t\in\mathcal{A}$.
\begin{equation}
\label{admissible_error_cond}
D_{\mathrm{TV}}(\bar{T}_{\theta,m}(\cdot|s_t,a_t),T(\cdot|s_t,a_t))\leq b\cdot\mathcal{U}^\mathrm{ADM}(s_t,a_t),
\end{equation}
where $\bar{T}_{\theta,m}$ is the overall conditioned distribution coming from
\begin{equation}
\label{overall_T}
\bar{T}_{\theta,m}(\cdot|s_t,a_t)=\frac{1}{m}\sum_{k=1}^m \left[\sum_{\substack{s_{t-k+1}\\a_{t-k+1:t-1}}}\Gamma^{k-1}_\pi(s_{t-k+1},a_{t-k+1:t-1}|s_t) \hat{T}_\theta(\cdot|s_{t-k+1},a_{t-k+1:t}))\right].
\end{equation}
\end{assumption}

Under Assumption \ref{admissible_error} and the $\xi$-uncertainty quantifier definition (see Appendix \ref{pevi_intro} for details) proposed by PEVI \cite{pevi}, we present the following theorem, demonstrating that $\mathcal{U}^{\mathrm{ADM}}$ can serve as a $\xi$-uncertainty quantifier to bound the Bellman error.
\begin{theorem}
\label{bellman_bound}
    $\beta\cdot \mathcal{U}^{\mathrm{ADM}}$ is a valid $\xi$-uncertainty quantifier, with $\beta=b\frac{\gamma r_{\mathrm{max}}}{1-\gamma}$. Specifically,
    \begin{equation}
        \left|\hat{\mathcal{T}}^\pi Q(s_t,a_t)-\mathcal{T}^\pi Q(s_t,a_t)\right|\leq\beta\cdot\mathcal{U}^{\mathrm{ADM}}(s_t,a_t),
    \end{equation}
    where $\hat{\mathcal{T}}^\pi$ is the proxy Bellman operator induced by ADM to estimate the true Bellman operator $\mathcal{T}^\pi$.
\end{theorem}
\begin{proof}
See Appendix \ref{theo}.
\end{proof}

According to the suboptimality theorem (see Appendix \ref{pevi_intro} for details) presented by PEVI \cite{pevi}, the policy $\hat{\pi}$ derived via pessimistic value iteration, which incorporates any $\xi$-uncertainty quantifier as a penalty term into the value iteration process \cite{rl}, has a bounded optimality gap to the optimal policy $\pi^*$. The optimality gap is dominated by the Bellman error and the uncertainty quantification. Intuitively, the Bellman error is usually small in safe regions where the dynamics model has been trained with rich data and tends to yield high consistency under different backtracking lengths, while large errors often appear in risky regions where data is scarce and the predictions via backtracking different lengths become inconsistent. The penalization prevents the policy from taking actions leading it to risky regions, otherwise the model will induce inaccurate value estimations on these actions. Thus, we can penalize the Bellman operator to obtain a pessimistic value estimation by
\begin{equation}
\label{ADM_bellman}
\hat{\mathcal{T}}^{\mathrm{ADM}}Q(s_t,a_t):=\hat{\mathcal{T}}^\pi Q(s_t,a_t)-\beta\cdot\mathcal{U}^{\mathrm{ADM}}(s_t,a_t).
\end{equation}

We expect the penalty term $\beta\cdot\mathcal{U}^{\mathrm{ADM}}(s_t,a_t)$ to be as small as possible thereby constraining the optimality gap. While our Assumption \ref{admissible_error} lacks theoretical guarantees and the tightness of the bound in Theorem \ref{bellman_bound} is unclear, we have provided sufficient evidence in Section \ref{u_quant} that our uncertainty quantification effectively estimates the model error.

Overall, ADMPO-OFF is the offline version of ADMPO-ON, which introduces a penalized Bellman operator \eqref{ADM_bellman} into the policy optimization process of ADMPO-ON, following the algorithmic framework of MOPO \cite{mopo}. The detailed pseudo-code is provided by Algorithm \ref{admpo_off_code} in Appendix \ref{admpo_off}.

\section{Experiments}
In this section, we conduct several experiments to answer: (1) Does ADM roll-out samples with less compounding error than the ensemble dynamics model? (2) How well does ADMPO-ON perform in the online setting? (3) How well does ADMPO-OFF perform in the offline setting? Does ADM quantify the model uncertainty better than the ensemble dynamics model?

\subsection{Dynamics Model Evaluation}
\begin{figure*}[pt!]
    \centering
    \includegraphics[width=\linewidth]{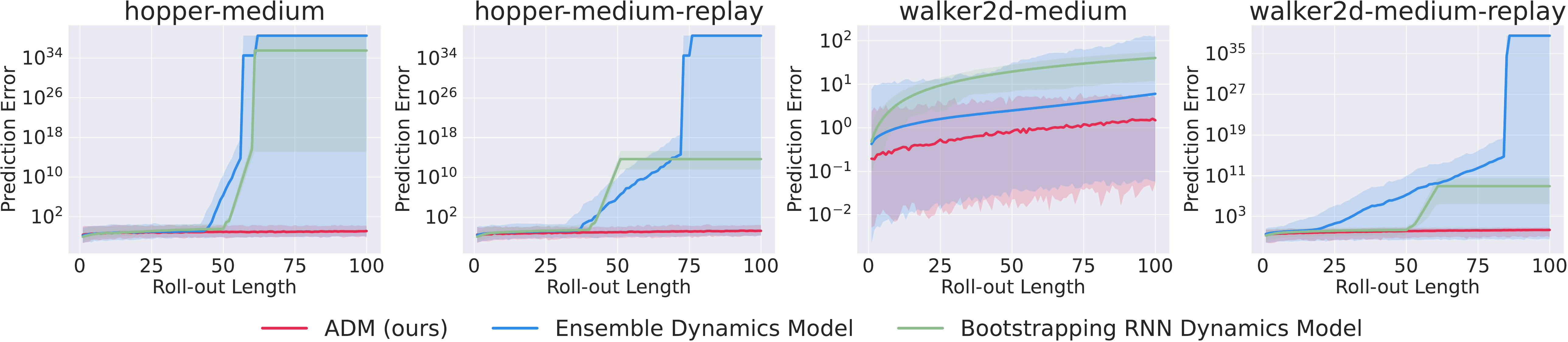}
    \caption{Comparison among ADM, ensemble dynamics model, and bootstrapping RNN dynamics model, in terms of the growth curve of the compounding error as roll-out length increases, after offline learning. The overflow value is regarded as the maximum value of \texttt{float32}.}
    \label{model_exp}
\end{figure*}

An essential metric for evaluating dynamics model quality is the compounding error, which increases with the roll-out length. We selected four D4RL \cite{d4rl} datasets, hopper-medium-v2, hopper-medium-replay-v2, walker2d-medium-v2, and walker2d-medium-replay-v2, to compare the compounding error between ADM and the commonly used ensemble dynamics model. To eliminate the influence of the RNN structure, we also compare the bootstrapping RNN dynamic model, which shares the same structure as ADM but predicts $s_{t+m+1}$ using the historical state-action sequence $(s_t,a_t,\cdots,s_{t+m},a_{t+m})$ as input. Figure \ref{model_exp} shows the growth curves of the compounding error as the roll-out length increases. We observe that the curves of ADM remain close to zero, while the other two models exhibit exponential growth as the roll-out length exceeds a certain threshold. This phenomenon suggests ADM can improve predictions for future states due to its any-step backtracking mechanism during model roll-outs.

\begin{figure*}[pt!]
    \centering
    \includegraphics[width=\linewidth]{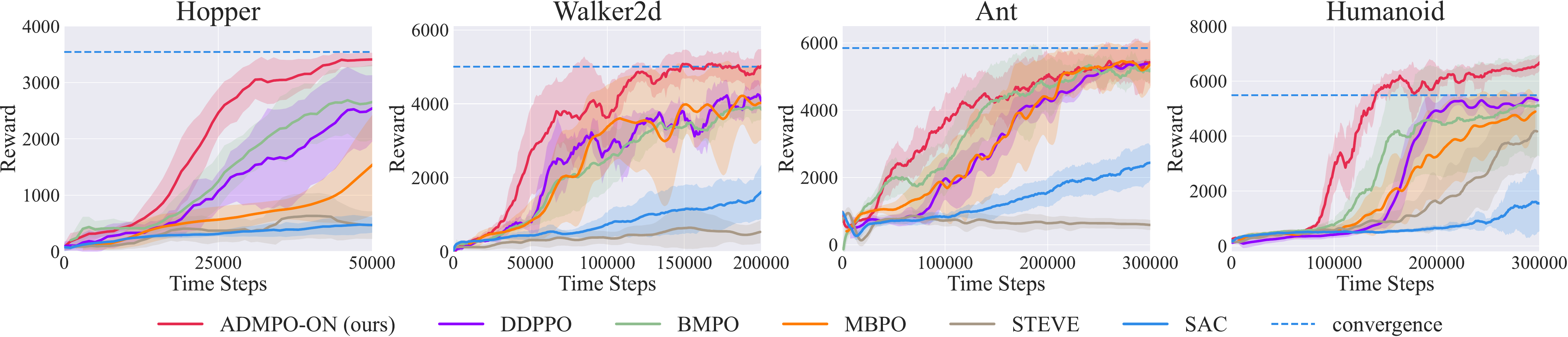}
    \caption{Online learning curves of ADMPO-ON (red) and other five baselines on four MuJoCo-v3 tasks. The blue dashed lines indicate the asymptotic performance of SAC for reference. The solid lines indicate the mean while the shaded areas indicate the standard error over five different seeds.}
    \label{mujoco_performance}
\end{figure*}

\subsection{Evaluation in Online Setting}
\label{online_evaluation}

We evaluate ADMPO-ON on four difficult MuJoCo continuous control tasks \cite{mujoco}, including Hopper, Walker2d, Ant, and Humanoid. All the tasks adopt version v3 and follow the default settings. Four model-based methods and one model-free method are selected as our baselines. These include SAC \cite{sac}, whcih is the state-of-the-art model-free RL algorithm; STEVE \cite{steve}, which incorporates an ensemble into the model-based value expansion; MBPO \cite{mbpo}, which updates the policy with a mixture of real environmental samples and branched roll-out data; BMPO \cite{bmpo}, which builds upon MBPO and replaces the dynamics model with a bidirectional one; and DDPPO \cite{ddppo}, which adopts a two-model-based learning method to control the prediction error and the gradient error.

Figure \ref{mujoco_performance} shows learning curves of ADMPO-ON and other five baselines, along with SAC's asymptotic performance. ADMPO-ON achieves competitive performance after fewer environmental steps than the baselines. Taking the most difficult Humanoid as an example, ADMPO-ON has achieved 100\% of SAC convergence performance (about 6000) after 150k steps, while DDPPO needs about 200k steps, and the other four methods can't get close to the blue dashed line even at step 300k. ADMPO-ON performs about 1.33x faster than DDPPO and dominates other baselines in terms of learning efficiency on the Humanoid task. After training, ADMPO-ON can achieve a final performance close to the asymptotic performance of SAC on all these four MuJoCo tasks. These results demonstrate that ADMPO-ON has both high sample efficiency and competitive performance. Further study on why ADMPO-ON performs well in the online setting can be found in Appendix \ref{sec_humanoid_exp}.

\subsection{Evaluation in Offline Setting}
\label{offline_evaluation}

\subsubsection{D4RL Benchmark Results}
We compare ADMPO-OFF with four model-free methods: BC (behavioral cloning), which simply imitates the behavior policy of the dataset; CQL \cite{cql}, which equally penalized the Q values on out-of-the-distribution state-action pairs; TD3+BC \cite{td3bc}, which simply incorporates a BC term into the policy optimization objective of TD3 \cite{td3}; and EDAC \cite{edac}, which quantifies the Q uncertainty via ensemble; as well as five model-based methods: MOPO \cite{mopo}, which adds the uncertainty of the model prediction as a penalization term to the reward function; COMBO \cite{combo}, which introduces the penalty function of CQL into the model-based framework; RAMBO \cite{rambo}, which adversarially trains the dynamics model and the policy; CBOP \cite{cbop}, which adopts the variance of values under an ensemble of dynamics models to estimate the Q value conservatively under MVE \cite{mve} regime; and MOBILE \cite{mobile}, which proposes Model-Bellman inconsistency to estimate the Bellman error.

Table \ref{d4rl_results} reports the results on twelve D4RL \cite{d4rl} MuJoCo datasets (v2 version). The normalized score for each dataset is obtained via online evaluation after offline learning. The source of the reported performance in provided in Appendix \ref{baseline_source}. We observe that ADMPO-OFF outperforms the other nine baselines in most tasks and achieves the highest average score.

\input{tables/d4rl}

\subsubsection{NeoRL Benchmark Results}
NeoRL \cite{neorl}, is an offline RL benchmark that collects the data in a manner more conservative and closer to real-world data-collection scenarios. We focus on nine datasets collected using policies of three different qualities (low, medium, and high) in three environments Hopper-v3, HalfCheetah-v3, and Walker2d-v3, respectively. In our evaluation, each dataset contains 1000 trajectories.

We compare our ADMPO-OFF with six baselines, including BC, CQL, TD3+BC, EDAC, MOPO, and MOBILE. Table \ref{neorl_results} presents the normalized scores of these methods. Due to the narrow and limited coverage of the NeoRL data, all the baselines experience a decline in performance. In contrast, our ADMPO-OFF maintains a relatively high-level average score, still achieving superior performance in most tasks. This remarkable out-performance indicates the potential of our algorithm in more challenging real-world tasks.

\input{tables/neorl}

\subsubsection{Uncertainty Quantification}
\label{u_quant}
\begin{figure*}[!t]
    \centering
    \subfigure[hopper-medium-replay-v2]{
        \includegraphics[width=0.48\linewidth]{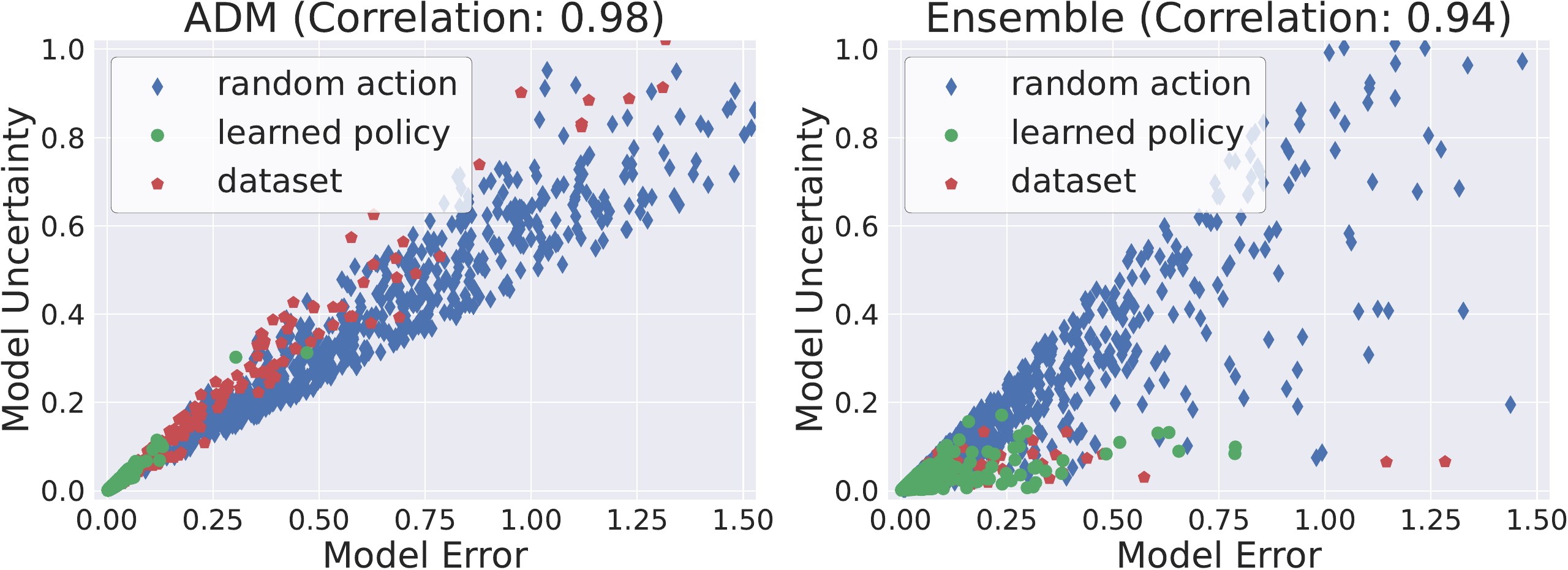}
    }
    \subfigure[walker2d-medium-replay-v2]{
        \includegraphics[width=0.48\linewidth]{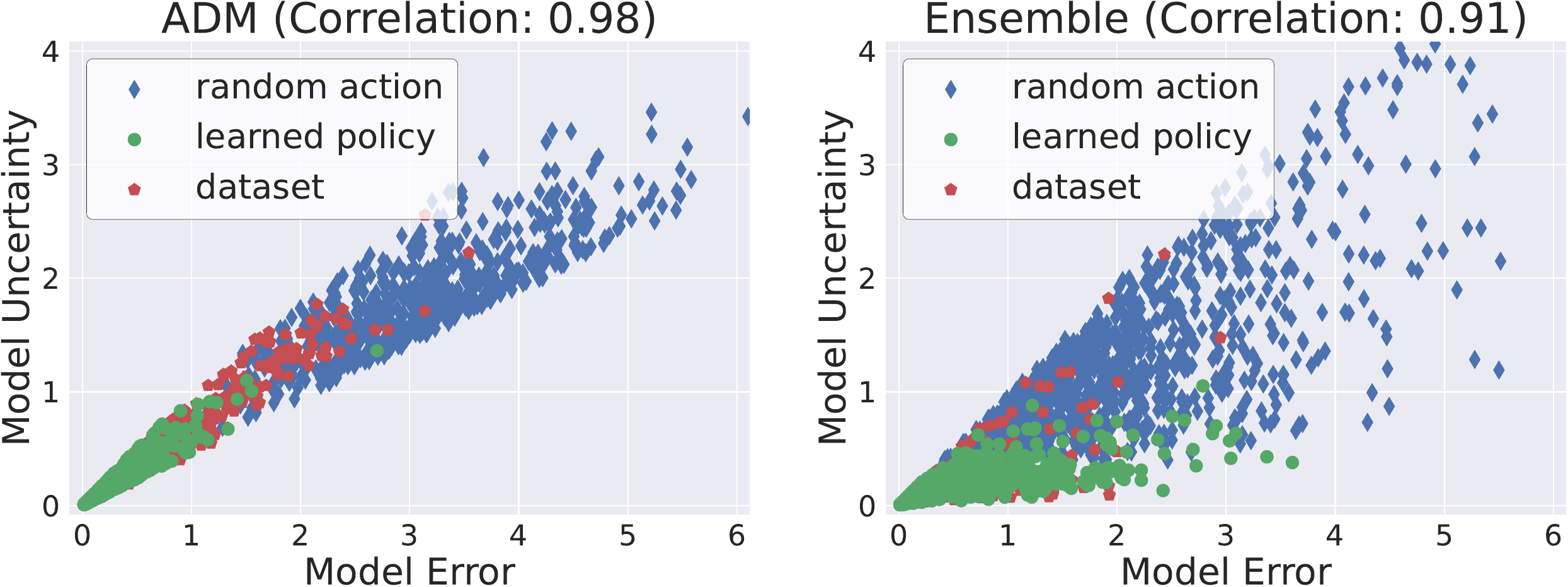}
    }
    \vspace{-3mm}
    \caption{Comparison between ADM and ensemble model in uncertainty quantification.}
    \label{uquant}
\end{figure*}

In our analysis, we sample a lot of state-action pairs in the learned ADM and the ensemble dynamics model respectively. These samples are obtained by model roll-out with three types of policy: random action selection, the learned policy after offline training, and the behavior policy of the dataset. Subsequently, we measure their model uncertainty and model error. The resulting scatter plots on two D4RL tasks, hopper-medium-replay-v2 and walker2d-medium-replay-v2, are illustrated in Figure \ref{uquant}. We observe that our ADM provides a better quantification for the model uncertainty. On the one hand, points sampled in ADM with greater model errors tend to exhibit greater quantified model uncertainty. The correlation coefficient of 0.98, observed across both tasks, surpasses that of the ensemble dynamics model. On the other hand, ADM can distinguish the samples from different policy better than the ensemble model. Samples generated from random actions deviate from the dataset distribution, whose uncertainty should be maximum in expectation. Conversely, when the learned policy is optimized within the safe regions covered by the dataset, model uncertainty is expected to be minimal. The experimental plots of ADM illustrate this phenomenon more clearly.

\section{Related Work}
This work is related to online and offline dyna-style MBRL \cite{dyna-q}.

\subsection{Online Model-based Reinforcement Learning}
In the online setting, MBRL algorithms aim to accelerate value estimation or policy optimization with model roll-out data. MVE \cite{mve} enhances Q-value target estimation by allowing short-term imagination to a fixed depth using the dynamics model. STEVE \cite{steve} builds upon MVE by incorporating an ensemble into the value expansion to better estimate the Q value. SLBO \cite{slbo} directly utilizes TRPO \cite{trpo} to optimize the policy with synthetic data generated by rolling out to the end of trajectories in the dynamics model. MBPO \cite{mbpo} proposes a branched roll-out scheme to truncate unreliable samples, thereby reducing the influence of compounding error \cite{xu2021error}, and employs SAC \cite{sac} to update the policy with a mixture of real-world data and model-generated data.

Recent work improves MBRL performance mainly from two perspectives. One focuses on learning a better dynamics model, such as bidirectional models \cite{bmpo}, adversarial models \cite{advm,armor}, causal models \cite{zhu2022arxiv}, and multi-step models \cite{msmbrl, m3, combining}. The other pursues a better utilization of the learned model, enhancing the reliability of model-generated samples \cite{m2ac} or applying model-based multi-step planning techniques \cite{pets, maac, dan, pinet, upn, mppve}.

\subsection{Offline Model-based Reinforcement Learning}
Although some model-free RL algorithms \cite{bear,bcq,td3bc,cql,edac,pbrl} have made significant contributions to offline RL research, MBRL algorithms appear to be more promising for the offline setting since they can utilize the dynamics model to extend the dataset and largely improve the data efficiency. 

The core issue of offline MBRL lies in how to effectively leverage the model. MOPO \cite{mopo} and MOReL \cite{morel} add the uncertainty of the model prediction as a penalization term to the original reward function to achieve a pessimistic value estimation. MOBILE \cite{mobile} improves the uncertainty quantification by introducing Model-Bellman inconsistency into the offline model-based framework. COMBO \cite{combo} applies CQL \cite{cql} to force the Q value to be small on model-generated out-of-distribution samples. RAMBO \cite{rambo} achieves conservatism by adversarial model learning for value minimization while keeping fitting the transition function. CBOP \cite{cbop} introduces adaptive weighting of short-horizon roll-out into MVE \cite{mve} technique and adopts the variance of values under an ensemble of dynamics models to estimate the Q value conservatively. MOREC \cite{morec} designs a reward-consistent dynamics model using an adversarial discriminator to let the model-generated samples be more reliable.

\section{Conclusion}
In this work, we propose a new method for environment model learning and utilization, namely \textbf{A}ny-step \textbf{D}ynamics \textbf{M}odel (ADM). ADM is applicable in both online and offline MBRL frameworks and yields two algorithms, ADMPO-ON and ADMPO-OFF, respectively. Several analysis and experiments show that ADM outperforms the ensemble dynamics model applied in previous MBRL approaches widely. The only problem is that RNN may consume more resources during the training process. We believe ADM has powerful potential beyond the capabilities demonstrated in this paper. In the future, we will explore the scalability of ADM in non-Markovian visual RL scenarios, considering both online and offline settings.

\normalem
\bibliographystyle{abbrv}
\bibliography{ref}


\newpage
\appendix

\section{Additional Introduction to Pessimistic Value Iteration (PEVI)}
\label{pevi_intro}
Pessimistic Value Iteration (PEVI) \cite{pevi} is a meta-algorithm for offline RL settings. It constructs an estimated Bellman operator $\hat{\mathcal{T}}^\pi$ based on the given dataset $\mathcal{D}_\mathrm{env}$ to approximate the true Bellman operator $\mathcal{T}^\pi$ that satisfies
\begin{equation}
    \mathcal{T}^\pi V_{h+1}(s_h,a_h)=\mathbb{E}_{(s_{h+1},r_{h+1})\sim T(\cdot|s_h,a_h)}\left[r_{h+1}+V_{h+1}(s_{h+1})\right],
\end{equation}
where $h$ is the step index less than the horizon $\mathcal{H}$. Then the state-action value function is updated with
\begin{equation}
    Q_h(s_h,a_h)\leftarrow \hat{\mathcal{T}}^\pi V_{h+1}(s_h,a_h)-\Lambda_h(s_h,a_h)
\end{equation}
for each $(s_h,a_h)$, where $\Lambda_h$ is the penalty function that guarantees the conservatism of the learned policy. Especially, $\Lambda_h$ should be a $\xi$-uncertainty quantifier as follows.
\begin{definition}[$\xi$-Uncertainty Quantifier (proposed by \cite{pevi})]
\label{xi_uquant}
The set of penalization $\{\Lambda_h\}_{h\in[\mathcal{H}]}$ forms a $\xi$-uncertainty quantifier if
\begin{equation}
\left|\hat{\mathcal{T}}^\pi V_{h+1}(s_h,a_h)-\mathcal{T}^\pi V_{h+1}(s_h,a_h)\right|\leq \Lambda_h(s_h,a_h)
\end{equation}
holds with probability at least $1-\xi$ for all $(s_h,a_h)\in\mathcal{S}\times\mathcal{A}$.
\end{definition}
The following theorem characterizes the suboptimality of PEVI.

\begin{theorem}[Suboptimality of PEVI (proposed by \cite{pevi})]
Suppose $\{\Lambda_h\}_{h=1}^\mathcal{H}$ in PEVI is a set of $\xi$-uncertainty quantifier. Then the derived policy $\hat{\pi}$ satisfies
\begin{equation}
    \left|V_1^{\pi^*}(s_1)-V_1^{\hat{\pi}}(s_1)\right|\leq2\sum_{h=1}^\mathcal{H}\mathbb{E}_{\rho^{\pi^*}}\left[\Lambda_h(s_h,a_h)\right]
\end{equation}
with probability at least $1-\xi$ for all starting $s_1\in\mathcal{S}$. Here $\mathbb{E}_{\rho^{\pi^*}}$ is with respect to the trajectory induced by the optimal policy $\pi^*$ in the underlying MDP given the fixed function $\Lambda_h$.
\end{theorem}
\begin{proof}
See PEVI \cite{pevi} for detailed proof.
\end{proof}

\section{Theoretical Results}
\label{theo}
\begin{theorem}
\label{bellman_bound2}
    $\beta\cdot \mathcal{U}^{\mathrm{ADM}}$ is a valid $\xi$-uncertainty quantifier, with $\beta=b\frac{\gamma r_{\mathrm{max}}}{1-\gamma}$. Specifically,
    \begin{equation}
        \left|\hat{\mathcal{T}}^\pi Q(s_t,a_t)-\mathcal{T}^\pi Q(s_t,a_t)\right|\leq\beta\cdot\mathcal{U}^{\mathrm{ADM}}(s_t,a_t),
    \end{equation}
    where $\hat{\mathcal{T}}^\pi$ is the proxy Bellman operator induced by ADM to estimate the true Bellman operator $\mathcal{T}^\pi$.
\end{theorem}
\begin{proof}
First, we define $y(\hat{s}_{t+1},\hat{r}_{t+1})=\hat{r}_{t+1}+\gamma \mathbb{E}_{a\sim\pi(\cdot|\hat{s}_{t+1})}\left[Q(\hat{s}_{t+1},a)\right]$ and expand these two Bellman operator to
\begin{equation}
\begin{aligned}
    &\hat{\mathcal{T}}^\pi Q(s_t,a_t)\\
    =&\mathbb{E}_{(s_{t-m+1:t-1},a_{t-m+1:t-1})\sim\Gamma^{m-1}_\pi(\cdot|s_t)}\left[\frac{1}{m}\sum_{k=1}^m\mathbb{E}_{(\hat{s}_{t+1},\hat{r}_{t+1})\sim \hat{T}_\theta(\cdot|s_{t-k+1},a_{t-k+1:t})}
\left[y(\hat{s}_{t+1},\hat{r}_{t+1})\right]\right] \\
=&\sum_{\substack{s_{t-m+1}\\a_{t-m+1:t-1}}}\Gamma^{m-1}_\pi(s_{t-m+1},a_{t-m+1:t-1}|s_t)\left[\frac{1}{m}\sum_{k=1}^m\sum_{\substack{\hat{s}_{t+1}\\\hat{r}_{t+1}}}\hat{T}_\theta(\cdot|s_{t-k+1},a_{t-k+1:t})y(\hat{s}_{t+1},\hat{r}_{t+1})\right]\\
=&\frac{1}{m}\sum_{k=1}^m \left[\sum_{\substack{s_{t-k+1}\\a_{t-k+1:t-1}}}\Gamma^{k-1}_\pi(s_{t-k+1},a_{t-k+1:t-1}|s_t) \sum_{\substack{\hat{s}_{t+1}\\\hat{r}_{t+1}}}\hat{T}_\theta(\cdot|s_{t-k+1},a_{t-k+1:t}))y(\hat{s}_{t+1},\hat{r}_{t+1})\right]\\
=&\sum_{\hat{s}_{t+1},\hat{r}_{t+1}}\bar{T}_{\theta,m}(\hat{s}_{t+1},\hat{r}_{t+1}|s_t,a_t)y(\hat{s}_{t+1},\hat{r}_{t+1}),
\end{aligned}
\end{equation}
and
\begin{equation}
\begin{aligned}
    &\mathcal{T}^\pi Q(s_t,a_t)\\
    =&\mathbb{E}_{\hat{s}_{t+1},\hat{r}_{t+1}\sim T(\cdot |s_t,a_t)}\left[y(\hat{s}_{t+1},\hat{r}_{t+1})\right]\\
    =&\sum_{\hat{s}_{t+1},\hat{r}_{t+1}}T(\hat{s}_{t+1},\hat{r}_{t+1}|s_t,a_t)y(\hat{s}_{t+1},\hat{r}_{t+1}).
\end{aligned}
\end{equation}
Then, we can obtain
\begin{equation}
\begin{aligned}
    &\left|\hat{\mathcal{T}}^\pi Q(s_t,a_t)-\mathcal{T}^\pi Q(s_t,a_t)\right|\\
    =&\sum_{\hat{s}_{t+1},\hat{r}_{t+1}}\left|\bar{T}_{\theta,m}(\hat{s}_{t+1},\hat{r}_{t+1}|s_t,a_t)-T(\hat{s}_{t+1},\hat{r}_{t+1}|s_t,a_t)\right|\cdot \left|y(\hat{s}_{t+1},\hat{r}_{t+1})\right|\\
    =&\gamma\sum_{\hat{s}_{t+1},\hat{r}_{t+1}}\left|\bar{T}_{\theta,m}(\hat{s}_{t+1},\hat{r}_{t+1}|s_t,a_t)-T(\hat{s}_{t+1},\hat{r}_{t+1}|s_t,a_t)\right|\cdot \left|\mathbb{E}_{a\sim\pi(\cdot|\hat{s}_{t+1})}\left[Q(\hat{s}_{t+1},a)\right]\right|\\
    \leq&\dfrac{\gamma r_{\mathrm{max}}}{1-\gamma}\sum_{\hat{s}_{t+1},\hat{r}_{t+1}}\left|\bar{T}_{\theta,m}(\hat{s}_{t+1},\hat{r}_{t+1}|s_t,a_t)-T(\hat{s}_{t+1},\hat{r}_{t+1}|s_t,a_t)\right|\\
    =&\dfrac{\gamma r_{\mathrm{max}}}{1-\gamma}D_{\mathrm{TV}}(\bar{T}_{\theta,m}(\cdot|s_t,a_t),T(\cdot|s_t,a_t))\\
    \leq&b\dfrac{\gamma r_{\mathrm{max}}}{1-\gamma}\mathcal{U}^{\mathrm{ADM}}(s_t,a_t).
\end{aligned}
\end{equation}
Thus, let $\beta=b\frac{\gamma r_{\mathrm{max}}}{1-\gamma}$,  we can say that $\beta\cdot \mathcal{U}^{\mathrm{ADM}}$ is a valid $\xi$-uncertainty quantifier, as defined by Definition \ref{xi_uquant}.
\end{proof}

\section{Implementation Details}

\subsection{ADMPO-ON}
\label{admpo_on}

Our ADMPO-ON algorithm follows the framework of MBPO \cite{mbpo}, as shown in Algorithm \ref{admpo_on_code}. The only difference between ADMPO-ON and MBPO lies in the way the dynamics model is trained and utilized, as indicated by the blue-highlighted parts in the pseudo-code.

\begin{algorithm}[h]
    \caption{ADMPO-ON}
    \label{admpo_on_code}

    \textbf{Input}: Initial ADM $\hat{T}_\theta$ and policy $\pi_\phi$, roll-out length $H$, maximum backtracking length $m$, real data buffer $\mathcal{D}_\mathrm{env}$, model data buffer $\mathcal{D}_\mathrm{model}$, wADM-up size $U$, interaction epochs $N$, steps per epoch $E$

    \begin{algorithmic}[1]
        \STATE Explore for $U$ environmental steps and add data to $\mathcal{D}_{\mathrm{env}}$
        \FOR{$N$ epochs}
            \STATE \textcolor{blue}{Train ADM $\hat{T}_\theta$ on $\mathcal{D}_\mathrm{env}$ by maximizing Equation \eqref{model_obj}}
            \FOR{$t=1$ to $E$}
                \STATE Sample action $a_t$ according to $\pi_\phi(\cdot|s_t)$
                \STATE Perform $a_t$ in the environment and add the real sample $(s_t,a_t,r_{t+1},s_{t+1})$ to $\mathcal{D}_\mathrm{env}$
                \FOR{$M$ model roll-outs}
                    \STATE \textcolor{blue}{Sample initial $m$-step state-action sequence $(s_{i:i+m-1},a_{i:i+m-2})$ from $\mathcal{D}_\mathrm{env}$}
                    \STATE \textcolor{blue}{Roll out $H$ steps in $\hat{T}_\theta$ via \textbf{ADM-Roll}($\hat{T}_\theta$, $\pi_\phi$, $H$, $m$, $(s_{i:i+m-1},a_{i:i+m-2})$) and add the model roll-out data to $\mathcal{D}_\mathrm{model}$}
                \ENDFOR
                \FOR{$G$ policy updates}
                    \STATE Update current policy $\pi_\phi$ using samples from $\mathcal{D}_{\mathrm{env}}\cup\mathcal{D}_{\mathrm{model}}$
                \ENDFOR
            \ENDFOR
        \ENDFOR
    \end{algorithmic}
\end{algorithm}

\subsection{ADMPO-OFF}
\label{admpo_off}

Our ADMPO-OFF algorithm follows the framework of MOPO \cite{mopo}, as shown in Algorithm \ref{admpo_off_code}. The only difference between ADMPO-OFF and MOPO lies lies in the way the dynamics model is trained and utilized, as indicated by the blue-highlighted parts in the pseudo-code.

\begin{algorithm}[h]
    \caption{ADMPO-OFF}
    \label{admpo_off_code}

    \textbf{Input}: Pre-collected dataset $\mathcal{D}_\mathrm{env}$, initial ADM $\hat{T}_\theta$ and policy $\pi_\phi$, roll-out length $H$, maximum backtracking length $m$, model data buffer $\mathcal{D}_\mathrm{model}$, iterations $N$, penalty coefficient $\beta$

    \begin{algorithmic}[1]
        \STATE \textcolor{blue}{Train ADM $\hat{T}_\theta$ on $\mathcal{D}_\mathrm{env}$ by maximizing Equation \eqref{model_obj}}
        \FOR{$N$ iterations}
            \FOR{$M$ model roll-outs}
                \STATE \textcolor{blue}{Sample initial $m$-step state-action sequence $(s_{i:i+m-1},a_{i:i+m-2})$ from $\mathcal{D}_\mathrm{env}$}
                \STATE \textcolor{blue}{Roll out $H$ steps in $\hat{T}_\theta$ via \textbf{ADM-Roll}($\hat{T}_\theta$, $\pi_\phi$, $H$, $m$, $(s_{i:i+m-1},a_{i:i+m-2})$)}
                \STATE \textcolor{blue}{Penalize the reward via $\tilde{r}=r-\beta\mathcal{U}^\mathrm{ADM}(s, a)$ for each rolled-out step}
                \STATE Add the penalized model roll-out data to $\mathcal{D}_\mathrm{model}$
            \ENDFOR
            \FOR{$G$ policy updates}
                \STATE Update current policy $\pi_\phi$ using samples from $\mathcal{D}_{\mathrm{env}}\cup\mathcal{D}_{\mathrm{model}}$
            \ENDFOR
        \ENDFOR
    \end{algorithmic}
\end{algorithm}

\subsection{Policy Optimization}
\label{policy optimization}
The policy optimization method used in our ADMPO-ON and ADMPO-OFF is SAC \cite{sac}, following MBPO \cite{mbpo} and MOPO \cite{mopo}. The hyper-parameters about SAC follow its standard implementation, as listed in Table \ref{sac_hyper}.
\input{tables/sac_hyper}

\section{Experimental Details}
\label{exp_details}
\subsection{Resource Requirements}
All experiments can be completed with just one NVIDIA GeForce RTX 2080 Ti or any other type of GPU with larger graphic memory. There are no additional resource requirements. The time of execution for each task is about 24 hours.

\subsection{ADMPO-ON Settings}
The experimental settings of our ADMPO-ON in Section \ref{online_evaluation} are listed in Table \ref{admpo_on_hyper}. 
\input{tables/admpo_on_hyper}

\subsection{ADMPO-OFF Settings}
The experimental settings of our ADMPO-OFF in Section \ref{offline_evaluation} are listed in Table \ref{admpo_off_hyper}.
\input{tables/admpo_off_hyper}

\newpage
\subsection{Source of Baselines' Results}
\label{baseline_source}

For the evaluation on D4RL \cite{d4rl} benchmarks, the results of the compared baselines come from two sources:
\begin{itemize}
    \item Retraining on D4RL datasets of v2 version with OfflineRL-Kit \cite{offinerlkit}, for the algorithms whose original papers only report the performance on the v0 version, such as CQL \cite{cql}, MOPO \cite{mopo}.
    \item Including the scores in their papers, for the algorithms whose original papers report the performance on the v2 version, such as TD3+BC \cite{td3bc}, EDAC \cite{edac}, RAMBO \cite{rambo}, CBOP \cite{cbop}, and MOBILE \cite{mobile}, or who does not provide source codes, such as COMBO \cite{combo}. 
\end{itemize}

For the evaluation on NeoRL \cite{neorl} benchmarks, we report the scores of BC, CQL, and MOPO from the original paper of NeoRL and retrain TD3+BC and EDAC with OfflineRL-Kit \cite{offinerlkit}.

\section{Additional Experiments}
\subsection{Study on Why ADMPO-ON Performs Well in Online Setting}
\label{sec_humanoid_exp}
\begin{figure*}[pt!]
    \centering
    \includegraphics[width=\linewidth]{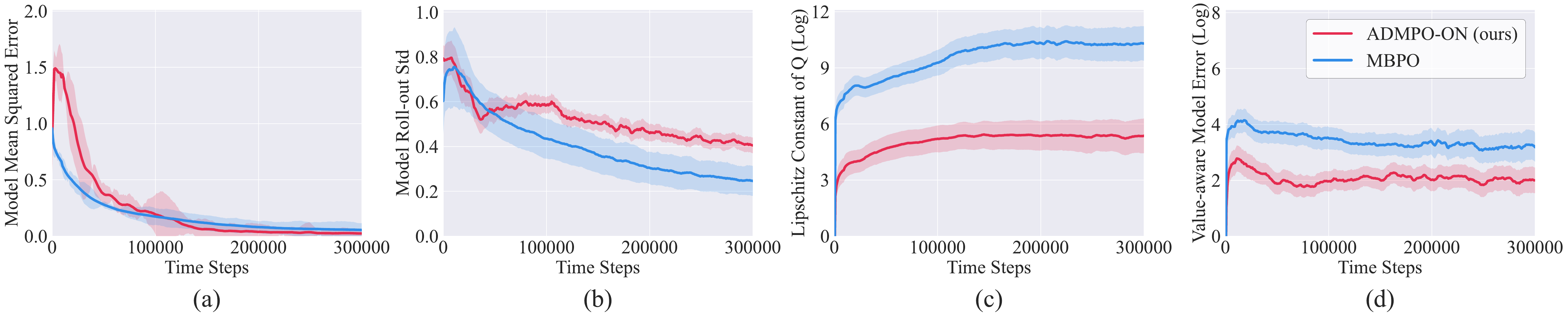}
    \caption{Comparison between ADMPO-ON and MBPO on Humanoid, in terms of (a) model mean squared error, (b) model roll-out standard deviation over diverse predictions, (c) estimated Lipschitz constant \cite{ensemble_necessary} of Q, and (d) value-aware model error \cite{value_aware_model_error}. Results are averaged over five seeds.}
    \label{humanoid_exp}
\end{figure*}

Value-aware model error \cite{value_aware_model_error} is a dependable metric for measuring the learning quality of the dynamics model and the suboptimality of the MBRL algorithm. We conduct a study to verify how well ADMPO-ON regulates the value-aware model error. Without loss of rigor, we only choose MBPO for comparison since most other model-based methods follow the same way of learning and utilizing the ensemble dynamics model. Figure \ref{humanoid_exp} shows the results on the most difficult Humanoid task. The learned ADM in ADMPO-ON and the ensemble dynamics model in MBPO achieve similar mean squared errors, indicating their similar fitting abilities. However, ADMPO-ON provides greater model roll-out standard deviation over diverse state prediction, forcing the agent to explore more uncertain areas. Therefore, since the variation of state prediction helps smoothen the Q target, the Q network in ADMPO-ON has a significantly smaller Lipschitz constant, and afterwards the value-aware model error, which measures the suboptimality of MBRL becomes smaller. This phenomenon explains why ADMPO-ON performs significantly better then MBPO in Figure \ref{mujoco_performance}. For details of the metrics used in this experiment, refer to \cite{ensemble_necessary}.

\subsection{Study on Maximum Backtracking Length}

\begin{figure*}[h!]
    \centering
    \includegraphics[width=\linewidth]{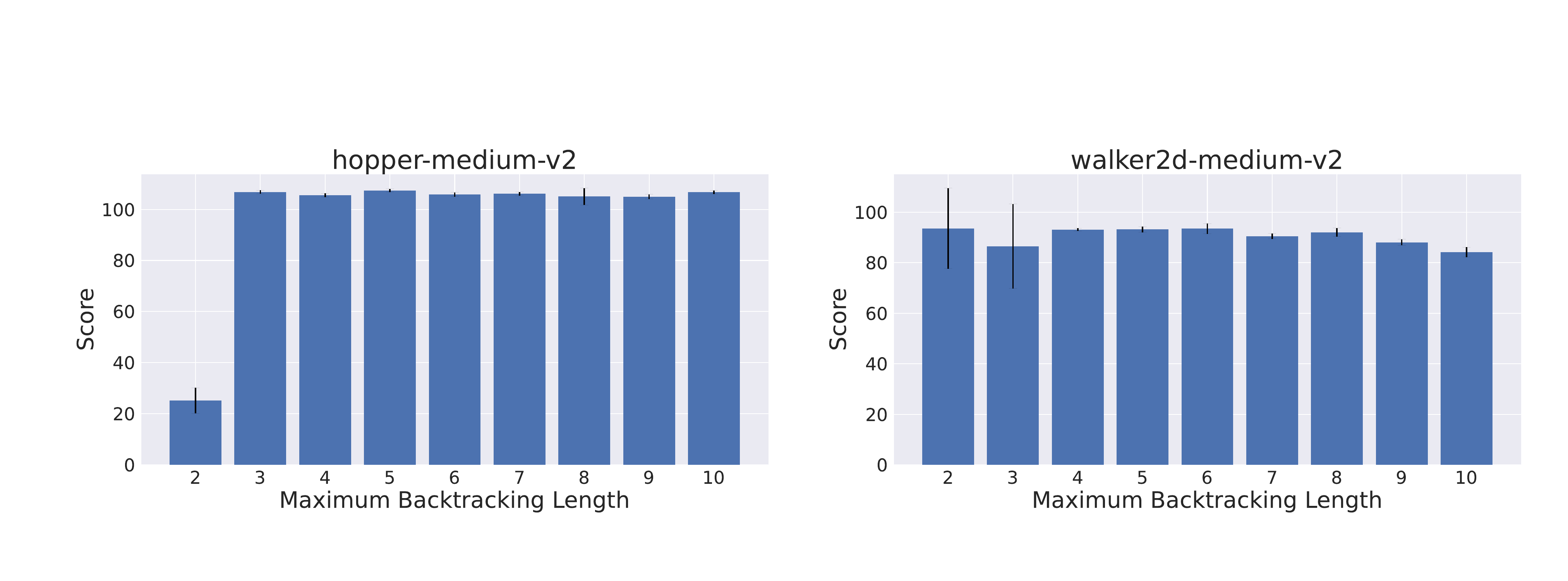}
    \caption{Illustration of ADMPO-OFF's performance under different maximum backtracking length.}
    \label{hyperm_exp}
\end{figure*}

Maximum backtracking length $m$ is an important hyper-parameter in our algorithms. Figure \ref{humanoid_exp} shows the influence of $m$ on the performance of hopper-medium-v2 and walker2d-medium-v2 tasks, respectively. We observe that the performance of ADMPO-OFF is not very sensitive to the hyper-parameter $m$.


\end{document}

%% file: tables/d4rl.tex
\begin{table*}[pt!]
	\centering
	\small
	\caption{Normalized scores after offline learning on D4RL MuJoCo tasks, averaged over five seeds.}
	\vspace{0.2em}
	\label{d4rl_results}
	\resizebox{\linewidth}{!}{
	\begin{tabular}{l|cccc|ccccc|c}
		\toprule
		\textbf{Task Name} & \textbf{BC} & \textbf{CQL} & \textbf{TD3+BC} & \textbf{EDAC} & \textbf{MOPO} & \textbf{COMBO} & \textbf{RAMBO} & \textbf{CBOP} & \textbf{MOBILE} & \textbf{ADMPO-OFF (ours)} \\
		\midrule
		hopper-random             & 3.7 & 5.3 & 8.5 & 25.3 & 31.7 & 17.9 & 25.4 & 31.4 & 31.9 & \textbf{32.7$\pm$0.2} \\
		halfcheetah-random        & 2.2 & 31.3 & 11.0 & 28.4 & 38.5 & 38.8 & 39.5 & 32.8 & 39.3 & \textbf{45.4$\pm$2.8} \\
		walker2d-random           & 1.3 & 5.4 & 1.6 & 16.6 & 7.4 & 7.0 & 0.0 & 17.8 & 17.9 & \textbf{22.2$\pm$0.2} \\
		\midrule
		hopper-medium             & 54.1 & 61.9 & 59.3 & 101.6 & 62.8 & 97.2 & 87.0 & 102.6 & 106.6 & \textbf{107.4$\pm$0.6} \\
		halfcheetah-medium        & 43.2 & 46.9 & 48.3 & 65.9 & 73.0 & 54.2  & \textbf{77.9} & 74.3 & 74.6 & 72.2$\pm$0.6 \\
		walker2d-medium           & 70.9 & 79.5 & 83.7 & 92.5 & 84.1 & 81.9 & 84.9 & \textbf{95.5} & 87.7 & 93.2$\pm$1.1 \\
		\midrule
		hopper-medium-replay      & 16.6 & 86.3 & 60.9 & 101.0 & 103.5 & 89.5 & 99.5 & \textbf{104.3} & 103.9 & \textbf{104.4$\pm$0.4} \\
		halfcheetah-medium-replay & 37.6 & 45.3 & 44.6 & 61.3  & \textbf{72.1} & 55.1  & 68.7 & 66.4 & 71.7 & 67.6$\pm$3.4 \\
		walker2d-medium-replay    & 20.3 & 76.8 & 81.8 & 87.1  & 85.6 & 56.0  & 89.2 & 92.7 & 89.9 & \textbf{95.6$\pm$2.1} \\
		\midrule
		hopper-medium-expert      & 53.9 & 96.9 & 98.0 & 110.7 & 81.6 & 111.1  & 88.2 & 111.6 & \textbf{112.6} & \textbf{112.7$\pm$0.3} \\
		halfcheetah-medium-expert & 44.0 & 95.0 & 90.7 & 106.3 & 90.8 & 90.0  & 95.4 & 105.4 & \textbf{108.2} & 103.7$\pm$0.2 \\
		walker2d-medium-expert    & 90.1 & 109.1 & 110.1 & 114.7 & 112.9 & 103.3  & 56.7 & \textbf{117.2} & 115.2 &  114.9$\pm$0.3 \\
		\midrule
		Average                   & 36.5 & 61.6 & 58.2 & 76.0 & 70.3 & 66.8  & 67.7 & 79.3 & 80.0 &  \textbf{81.0} \\
		\bottomrule
	\end{tabular}
	}
\end{table*}

%% file: tables/neorl.tex
\begin{table*}[!pt]
	\centering
    \scriptsize
	\caption{Normalized scores after offline learning on NeoRL tasks, averaged over five seeds.}
	\vspace{0.2em}
	\label{neorl_results}
	\begin{tabular}{l|cccccc|c}
		\toprule
		\textbf{Task Name} & \textbf{BC} & \textbf{CQL} & \textbf{TD3+BC} & \textbf{EDAC} & \textbf{MOPO} & \textbf{MOBILE} & \textbf{ADMPO-OFF (ours)} \\
		\midrule
		neorl-hopper-low & 15.1 & 16.0 & 15.8 & 18.3 & 6.2 & 17.4 & \textbf{22.3$\pm$0.1} \\
		neorl-halfcheetah-low & 29.1 & 38.2 & 30.0 & 31.3 & 40.1 & \textbf{54.7} & 52.8$\pm$1.2 \\
		neorl-walker2d-low & 28.5 & 44.7 & 43.0 & 40.2 & 11.6 & 37.6 & \textbf{55.9$\pm$3.8} \\
	    \midrule
		neorl-hopper-medium & 51.3 & 64.5 & \textbf{70.3} & 44.9 & 1.0 & 51.1 & 51.5$\pm$5.0 \\
		neorl-halfcheetah-medium & 49.0 & 54.6 & 52.3 & 54.9 & 62.3 & \textbf{77.8} & 69.3$\pm$1.7 \\
		neorl-walker2d-medium & 48.7 & 57.3 & 58.5 & 57.6 & 39.9 & 62.2 & \textbf{70.1$\pm$2.4} \\
		\midrule
		neorl-hopper-high & 43.1 & 76.6 & 75.3 & 52.5 & 11.5 & \textbf{87.8} & \textbf{87.6$\pm$4.9} \\
		neorl-halfcheetah-high & 71.3 & 77.4 & 75.3 & 81.4 & 65.9 & 83.0 & \textbf{84.0$\pm$0.8} \\
		neorl-walker2d-high & 72.6 & 75.3 & 69.6 & 75.5 & 18.0 & 74.9 & \textbf{82.2$\pm$1.9} \\
		\midrule
		Average & 45.4 & 56.1 & 54.5 & 50.7 & 28.5 & 60.7 & \textbf{64.0} \\
		\bottomrule
	\end{tabular}
\end{table*}

%% file: tables/sac_hyper.tex
\begin{table}[h]
	\centering
	\small
	\caption{Hyper-parameters of Policy Optimization in ADMPO-ON and ADMPO-OFF.}
	\vspace{0.2em}
	\label{sac_hyper}
	\begin{tabular}{l l l}
		\toprule
		\textbf{Hyper-parameter} & \textbf{Value} & \textbf{Description} \\
		\midrule
		$N_Q$ & 2 & the number of critics. \\
		actor network & FC(256,256) & fully connected (FC) layers with ReLU activations. \\
		critic network & FC(256,256) & fully connected (FC) layers with ReLU activations. \\
		$\tau$ & $5\times 10^{-3}$ & target network smoothing coefficient. \\
		$\gamma$ & 0.99 & discount factor. \\
		$lr_{\textrm{actor}}$ & $1\times 10^{-4}$ & learning rate of actor. \\
		$lr_{\textrm{critic}}$ & $3\times 10^{-4}$ & learning rate of critic. \\
		optimizer & Adam & optimizers of the actor and critics. \\
		batch size & 256 & batch size for each update. \\
		\bottomrule
	\end{tabular}
\end{table}

%% file: tables/admpo_on_hyper.tex
\begin{table*}[h!]
\centering
\small
\caption{Hyper-parameter settings of ADMPO-ON results presented in Figure \ref{mujoco_performance}. $x\rightarrow y$ over $a\rightarrow b$ denotes a thresholded linear increasing schedule, \textit{i.e.} the length of model rollouts at step $t$ is calculated by $f(t)=\min \left(\max \left(x+\frac{t-a}{b-a} \cdot(y-x), x\right), y\right)$.}
\begin{tabular}{|c|c|c|c|c|}
\hline
\multirow{2}{*}{\textbf{environment}}  & \multirow{2}{*}{Hopper}  & \multirow{2}{*}{Walker2d} & \multirow{2}{*}{Ant} & \multirow{2}{*}{Humanoid}\\
\multirow{2}{*}{}  & \multirow{2}{*}{}  & \multirow{2}{*}{}  & \multirow{2}{*}{} & \multirow{2}{*}{}\\
\hline
\multirow{2}{*}{\textbf{steps}}  & \multirow{2}{*}{50k} & \multirow{2}{*}{200k} & \multicolumn{2}{c|}{\multirow{2}{*}{300k}} \\
\multirow{2}{*}{} & \multirow{2}{*}{}  & \multirow{2}{*}{} & \multicolumn{2}{c|}{\multirow{2}{*}{}} \\

\hline
\multirow{2}{*}{\textbf{Update-To-Date ratio}} & \multicolumn{4}{c|}{\multirow{2}{*}{20}} \\
\multirow{2}{*}{} & \multicolumn{4}{c|}{\multirow{2}{*}{}}\\

\hline
\multirow{2}{*}{\textbf{maximum backtracking length} $\boldsymbol{m}$} & \multirow{2}{*}{5} & \multicolumn{3}{c|}{\multirow{2}{*}{2}} \\
\multirow{2}{*}{}& \multirow{2}{*}{} & \multicolumn{3}{c|}{\multirow{2}{*}{}}\\

\hline
\multirow{2}{*}{\textbf{model rollout schedule}}  & 1$\rightarrow$15 over  & 1$\rightarrow$10 over  & 1$\rightarrow$5 over & 1$\rightarrow$10 over \\
\multirow{2}{*}{} & 0$\rightarrow$50k & 0$\rightarrow$100k  &  10k$\rightarrow$100k &  10k$\rightarrow$100k\\

\hline
\multirow{2}{*}{\textbf{target entropy}}  & \multirow{2}{*}{-1} & \multirow{2}{*}{-3} & \multirow{2}{*}{-4} & \multirow{2}{*}{-8}\\
\multirow{2}{*}{} & \multirow{2}{*}{} & \multirow{2}{*}{} & \multirow{2}{*}{} & \multirow{2}{*}{} \\
\hline

\end{tabular}
\label{admpo_on_hyper}
\end{table*}

%% file: tables/admpo_off_hyper.tex
\begin{table}[ht]
        \linespread{0.8}
	\centering
	\small
	\caption{Hyper-parameter settings of ADMPO-OFF results presented in Section \ref{offline_evaluation}.}
	\vspace{0.2em}
	\label{admpo_off_hyper}
	\begin{tabular}{l | c c c c}
		\toprule
		\textbf{Domain Name} & \textbf{Task Name} & $\boldsymbol{m}$ & $\boldsymbol{H}$ & $\boldsymbol{\beta}$\\
            \midrule
            \multirow{12}{*}{D4RL MuJoCo}
		& hopper-random & 5 & 50 & 5 \\
		& halfcheetah-random & 2 & 10 & 2.5 \\
		& walker2d-random & 2 & 50 & 2.5 \\
		\cmidrule{2-5}
		& hopper-medium & 5 & 10 & 1 \\
		& halfcheetah-medium & 2 & 5 & 2.5 \\
		& walker2d-medium & 5 & 10 & 5 \\
		\cmidrule{2-5}
		& hopper-medium-replay & 5 & 5 & 0.1 \\
		& halfcheetah-medium-replay & 2 & 5 & 2.5 \\
		& walker2d-medium-replay & 5 & 5 & 0.1 \\
		\cmidrule{2-5}
		& hopper-medium-expert & 2 & 20 & 20 \\
		& halfcheetah-medium-expert & 2 & 50 & 10 \\
		& walker2d-medium-expert & 3 & 2 & 6 \\
		\midrule
  
        \multirow{9}{*}{NeoRL MuJoCo}
		& neorl-hopper-low & 5 & 20 & 5 \\
		& neorl-halfcheetah-low & 2 & 20 & 10 \\
		& neorl-walker2d-low & 5 & 10 & 2.5 \\
		\cmidrule{2-5}
		& neorl-hopper-medium & 5 & 20 & 50 \\
		& neorl-halfcheetah-medium & 2 & 5 & 20 \\
		& neorl-Walker2d-medium & 5 & 10 & 5 \\
		\cmidrule{2-5}
		& neorl-hopper-high & 5 & 20 & 50 \\
		& neorl-halfcheetah-high & 2 & 10 & 50 \\
		& neorl-walker2d-high & 5 & 10 & 2.5 \\
		\bottomrule
	\end{tabular}
\end{table}